\documentclass[preprint,12pt]{elsarticle}

\usepackage{amssymb}
\usepackage{times}
\usepackage{epsfig}
\usepackage{graphicx}
\usepackage{amsmath}
\usepackage{caption}
\usepackage{subcaption}

\journal{Pattern Recognition Letters}

\begin{document}

\begin{frontmatter}



\title{Multispectral Spatial Characterization: Application to Mitosis Detection in Breast Cancer Histopathology}

 \author[ujf,ipal]{H.~Irshad\corref{cor}}
 \author[cosmo,temasys]{A.~Gouaillard}
 \author[ujf,ipal]{L.~Roux}
 \author[upmc,ipal]{D.~Racoceanu}

 \cortext[cor]{Corresponding author}

 \address[ujf]{University Joseph Fourier, Grenoble France}
 \address[umpc]{University Pierre and Marie Curie, Paris, France}
 \address[ipal]{IPAL, CNRS UMI 2955, Singapore}
 \address[cosmo]{CoSMo Software, Boston, MA, USA}
 \address[temasys]{Temasys Communications, Singapore}

\author{}

\address{}

\begin{abstract}
Accurate detection of mitosis plays a critical role in breast cancer histopathology. Manual detection and counting of mitosis is tedious and subject to considerable inter- and intra-reader variations. Multispectral imaging is a recent medical imaging technology, proven successful in increasing the segmentation accuracy in other fields. This study aims at improving the accuracy of mitosis detection by developing a specific solution using multispectral and multifocal imaging of breast cancer histopathological data. We propose to enable clinical routine-compliant quality of mitosis discrimination from other objects. The proposed framework includes comprehensive analysis of spectral bands and z-stack focus planes, detection of expected mitotic regions (candidates) in selected focus planes and spectral bands, computation of multispectral spatial features for each candidate, selection of multispectral spatial features and a study of different state-of-the-art classification methods for candidates classification as mitotic or non mitotic figures. This framework has been evaluated on MITOS multispectral medical dataset and achieved 60\% detection rate and 57\% F-Measure. Our results indicate that multispectral spatial features have more information for mitosis classification in comparison with white spectral band features, being therefore a very promising exploration area to improve the quality of the diagnosis assistance in histopathology.
\end{abstract}


\end{frontmatter}

%%%%%%%%% BODY TEXT
\section{Introduction}
Breast Cancer (BC) is the most commonly diagnosed cancer after the skin cancers, and the second leading cause of cancer death, following lung cancer among U.S. women \cite{jiemin2013}. In 2012, an estimated 226,870 new cases of invasive BC and 39,510 BC deaths in women are reported in U.S. In addition, BC incidence and mortality rates have been increasing rapidly in economically less developed countries. According to World Health Organization, the reference process for breast cancer prognosis is histologic grading that combine tubule formation, nuclei atypia and mitotic counts \cite{bloom1957, elston1993}. This assessment of tissue sample is synthesized into a diagnosis that would help the clinician determine the best course of therapy. Several CAD solutions exist for the detection of tubule formation \cite{petushi2006, naik2008} and nuclei atypia \cite{cosatto2008, dalle2009, chaudhury2011, dundar2011} but only few are dedicated to mitotic counts (MC) \cite{irshad2013a, irshad2013b}.
 
In histopathology, H\&E is a well-established staining technique that exploits intensity of stains in the tissue images to quantify the nuclei and other structures related to cancer developments \cite{avwioro2011}. In this context, image-processing techniques are devoted to accurate and objective quantification and localization of cancer evolution in specific regions of the tissue such as cytoplasm, membranes and nuclei \cite{meijer1997}. From the chromatic viewpoint, nuclear regions are characterized by non-uniform stain intensity and color, thus preventing a trivial classification based on color separation. In addition, the superposition of tissue layers, as well as the diffusion of the dyes on the tissue surface, may bring the stains to contaminate the background or other cellular regions, which are different from their specific target. 

One of the most difficult fields in histopathological dataset analysis is spatial analysis, more specifically automated nuclei detection and classification \cite{fuchs2011}. The objective of nuclei classification is to assign different labels to different type of nuclei as normal, cancer, mitotic, apoptosis, lymphocytes etc. In addition, quantitative characterization is important not only for clinical applications (e.g., to reduce/eliminate inter- and intra-observer variation in diagnosis) but also for research applications (e.g., to understand the biological mechanisms of the disease process) \cite{gurcan2009}.

Image analysis in cytology has been studied for years and numerous solutions \cite{wolberg1993, stewart1998, cibas2009, plissiti2013, gong2013} have thus been proposed in the literature. The application of these solutions to histopathology is rather complicated due to the radical differences between the two imaging modalities and to the highly complex image characteristics. Indeed, in the case of histology images, cellular structures and functions are studied embedded in the whole tissue structure, presenting various cells architecture (gland formation, DCIS) very difficult to handle with usual pattern recognition techniques. Nevertheless, recent work \cite{malon2013, irshad2013a, irshad2013b} show great potential for Computer Assisted Diagnostic of histopathological datasets for breast cancer diagnosis.
 
Multispectral imaging (MSI) has the advantage to retrieve spectrally resolved information of a tissue image scene at specific frequencies across the electromagnetic spectrum. MSI captures images with accurate spectral content correlated with spatial information and reveals the chemical and anatomic features of histopathology \cite{levenson2006b, levenson2008}. This modality provides option to biologists and pathologists to see beyond the RGB image planes that they are accustomed to. Recent publications \cite{fernandez2005, levenson2006, wu2012, khelifi2012} have begun to explore the use of extra information contained in such spectral data. Specifically, there have been comparisons of spectral methodologies which demonstrate the advantage of multispectral data \cite{levenson2003, gentry1999}. The added benefit of MSI for analysis in routine H\&E histopathology, however, is still largely unknown, although some promising results are presented in \cite{roula2003, fernandez2005, khelifi2012, wu2012}. As far as we know, there is no existing study of the advantage, or lack of thereof, of MSI for automation of MCs in breast cancer histopathology. We propose here to extend the already successful work of \cite{irshad2013b} to support MSI and illustrate the advantages, paving the way for a better, automated detection of MCs in BC histopathology.

The reminder of the paper is organized as follows. Section~\ref{sec:previous} reviews the state-of-the-art multispectral methods, particularly in object or region detection in histopathology, related to this research work. Section~\ref{sec:framework} describes the proposed framework for mitotic figure detection. Experiment and results are presented in section~\ref{sec:results}. Section~\ref{sec:Discussion} contain the discussion part. Finally, the concluding remarks with future work are given in section~\ref{sec:conclusion}.

%-------------------------------------------------------------------------
\section{Literature Review} \label{sec:previous}
The main idea for extracting texture features from MSI is the use of combined spectral and spatial information for discrimination of regions or objects. We found few methods in the MSI literature for texture characterization of histopathological images. Some of them employed single band of MSI \cite{masood2009,wu2012,malon2013} and other used multiple bands of MSI \cite{khelifi2012, boucheron2007}. Fernandez et al. \cite{fernandez2005} coupled high-throughput Fourier transform infra-red (FTIR) spectroscopic imaging of tissue microarrays with statistical pattern recognition of spectra indicative of endogenous molecular composition and demonstrate histopathological characterization of prostatic tissue. They explicitly defined metrics consisting of spectral features that have a physical significance related to tissue biochemistry and facilitating the measurement of cell types. 

Recently, Masood et al. \cite{masood2009} proposed a colon biopsy classification method based on spatial analysis of hyper-spectral image data from colon biopsy samples. Initially, using circular local binary pattern algorithm, spatial analysis of patterns is represented by a feature vector in selected spectral band. Later, classification is achieved using subspace projection methods like principal component analysis, linear component analysis and support vector machine. Recently, Wu et al. \cite{wu2012} proposed a multilayer conditional random field model using a combination of low-level cues and high-level contextual information for nuclei separation in high dimensional data set obtained through spectral microscopy. In this approach, the multilayer contextual information was extracted by an unsupervised topic discovery process from spectral images of microscopic specimen, which efficiently helps to suppress segmentation errors caused by intensity inhomogeneity and variable chromatin texture. Malon et al. \cite{malon2013} demonstrated a segmentation based features with convolutional neural networks using selected focus plane and spectral band for identification of mitotic figures in breast cancer histopathology and achieved best classification accuracy (F-measure = 59\%) on multispectral dataset during ICPR contest 2012 \cite{roux2013}. These approaches \cite{masood2009, wu2012, malon2013} are limited to single spectral band by ignoring additional relevant information from other spectral bands.

Boucheron et al. \cite{boucheron2007} presented an analysis that the additional spectral bands contain further information useful for nuclear classification in histopathology as compared to the three standard bands of RGB imagery. Using all image bands, they reported $0.79\%$ improved in performance as compared to next best performing image type. This method is limited to pixel level features. One possibility of additional improvement in object classification is multispectral spatial analysis using texture features.

Khelifi et al. \cite{khelifi2012} proposed a spatial and spectral gray level dependence method in order to extend the concept of gray level co-occurrence matrix by assuming the presence of texture joint information between spectral bands. Having the fact that some spectral bands have more relevant information for specific object or region classification as compared to other spectral bands. This approach is limited to single spatial feature as computed from all spectral bands rather than one different spatial feature for each spectral band.

Each object or region has different level of relevant information. In the proposed methodology, we address the shortcomings of previous works, including (1) comprehensive analysis of multispectral spatial features (MSSF) in all bands rather than single band \cite{masood2009,wu2012,malon2013} and (2) extracting MSSF in order to discriminate mitotic figures from other nuclei and microscopic objects. The main novel contributions of proposed work are (1) selection of z-stack focus planes and multispectral bands for detection of candidates; (2) a multispectral spatial and morphology features computation which leverages discriminant information from a given nucleus across bands and (3) a selection of features from multispectral features vector for classification of mitotic figures in breast cancer histopathological images.

%------------------------------------------------------------------------
\section{Proposed Framework} \label{sec:framework}
\subsection{Dataset}
\begin{figure}[t]
	\centering
	\includegraphics[width=\textwidth]{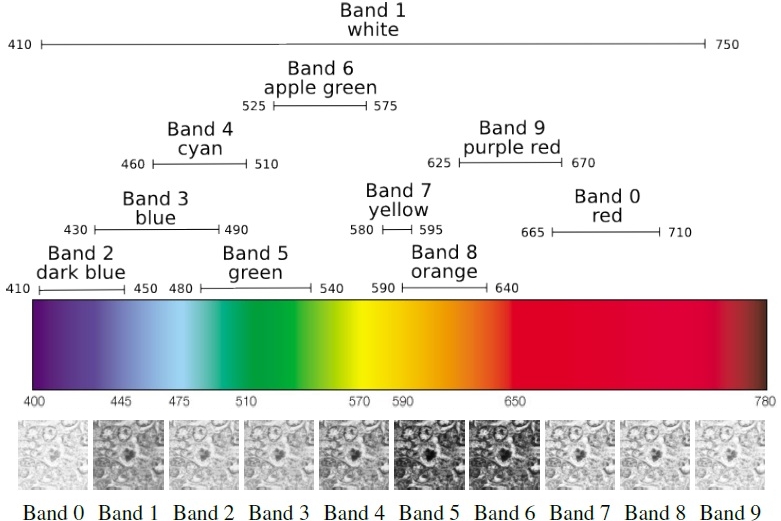}
	\caption{Spectral bands of the multispectral microscope and examples for each band.}
	\label{fig:spectral_bands}	
\end{figure}

We evaluated the proposed framework on multispectral MITOS dataset \cite{mITOS2012} a freely available medical mitosis dataset. The data set is made up of 50 high power fields (HPF) coming from five different slides scanned at 40X magnification using a 10 bands multispectral microscope. There are 10 HPFs per slide and each HPF has a size of $512\times512\mu\text{m}^2$ (that is an area of 0.262 mm\textsuperscript{2}). The spectral bands are all in the visible spectrum. In addition, for each spectral band, the digitization has been performed at 17 different focus planes (17 layers Z-stack), each plane being separated from the other by 500 nm. For one HPF, there are 170 gray scale images (10 spectral bands and 17 layers Z-stack for each spectral band). These 50~HPFs contain a total 322 mitotic figures. The training data set consists of 35~HPFs containing 224 mitotic figures and evaluation data set consists of 15~HPFs containing 98 mitotic figures \cite{roux2013}. Fig.~\ref{fig:spectral_bands} shows the spectral coverage of each of the 10~spectral bands of the multispectral microscope.

\subsection{Proposed Framework}
\begin{figure}
	\centering
	\includegraphics[width=\textwidth,height=40mm]{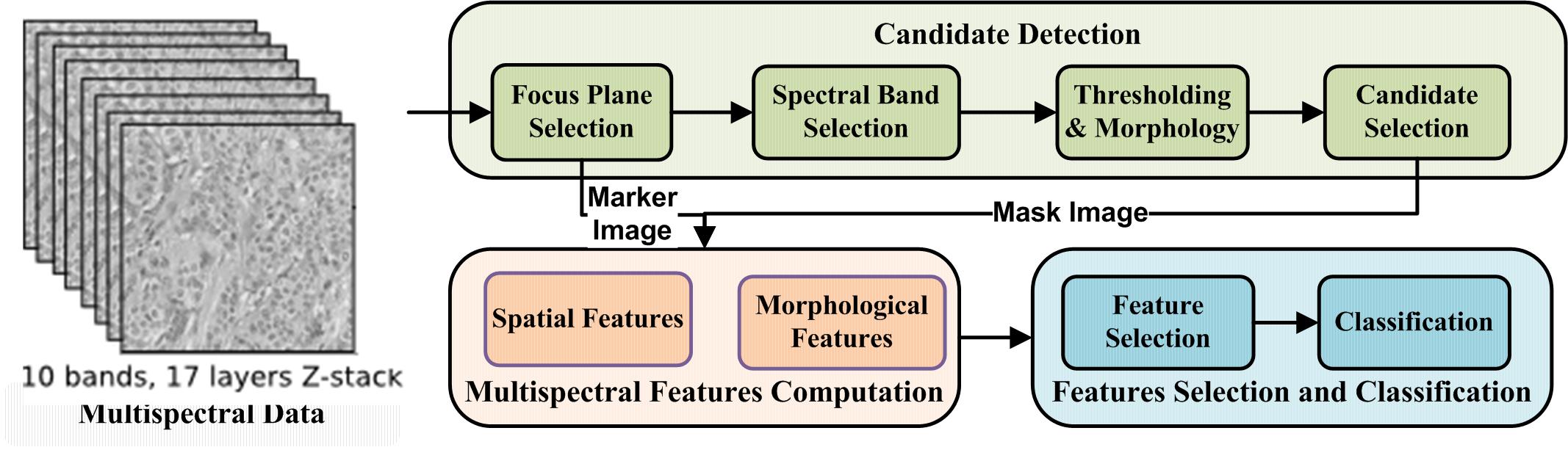}
	\caption{Proposed Framework.}
	\label{fig:framework}
\end{figure}

In this paper, we propose a framework for MC in Multispectral BC histopathology as shown in Fig.~\ref{fig:framework}. A comprehensive analysis of all multispectral bands (10~bands) and z-stack focus planes is performed for detection of mitosis candidates. Then candidates are detected using thresholding and morphological operations on selected band and focus plane. A multispectral features vector is computed for detected candidates having intensity and texture features across all bands of multispectral images. In addition, using segmented regions of detected candidates, morphological features are also computed. A feature selection algorithm is employed on this multispectral features vector in order to save the computation cost, to discard any redundancy in the data, and to improve classification accuracy. Classification is achieved using Bayesian, decision tree (DT), neural network as well as linear and non-linear support vector machine (SVM) classifiers. A side advantage of performing the spatial analysis on a multiple band simultaneously is to investigate whether improvement in accuracy can be achieved with carefully selected multispectral features over those methods \cite{masood2009,wu2012,malon2013} which use single band data.

\subsubsection{Candidate Detection}
\begin{figure}
	\centering
	\begin{subfigure}[b]{0.30\textwidth}
		\includegraphics[width=\textwidth]{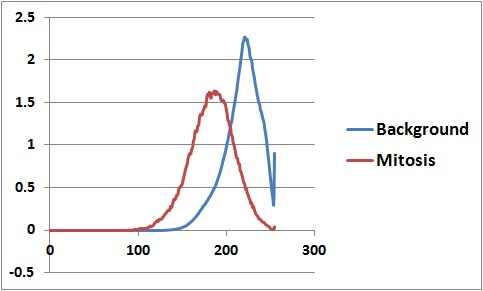}
		\caption*{a. Spectral Band 0}
	\end{subfigure}
	\begin{subfigure}[b]{0.30\textwidth}
		\includegraphics[width=\textwidth]{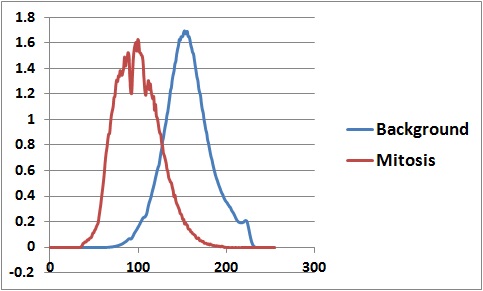}
		\caption*{b. Spectral Band 1}
	\end{subfigure}
	\begin{subfigure}[b]{0.30\textwidth}
		\includegraphics[width=\textwidth]{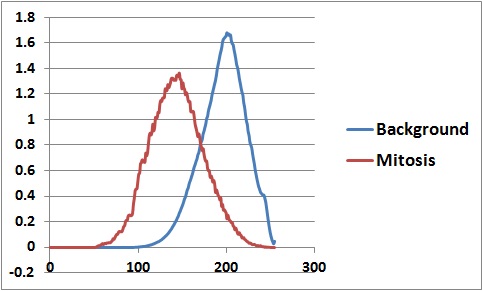}
		\caption*{c. Spectral Band 2}
	\end{subfigure}
	\begin{subfigure}[b]{0.30\textwidth}
		\includegraphics[width=\textwidth]{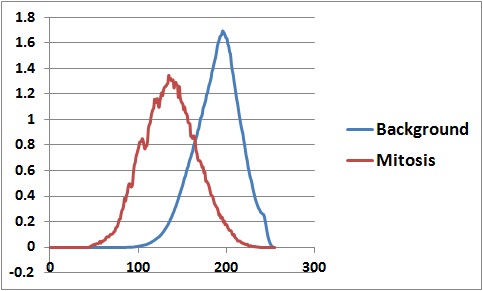}
		\caption*{d. Spectral Band 3}
	\end{subfigure}
	\begin{subfigure}[b]{0.30\textwidth}
		\includegraphics[width=\textwidth]{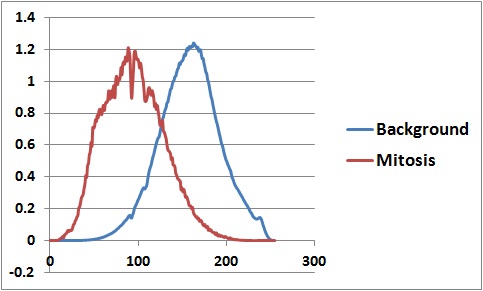}
		\caption*{e. Spectral Band 4}
	\end{subfigure}
	\begin{subfigure}[b]{0.30\textwidth}
		\includegraphics[width=\textwidth]{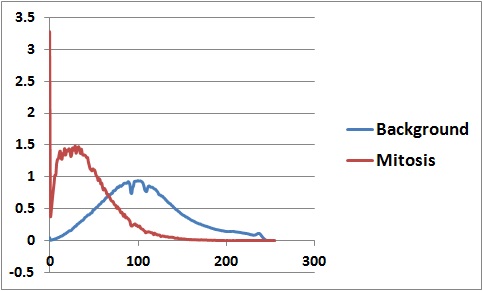}
		\caption*{f. Spectral Band 5}
	\end{subfigure}
	\begin{subfigure}[b]{0.30\textwidth}
		\includegraphics[width=\textwidth]{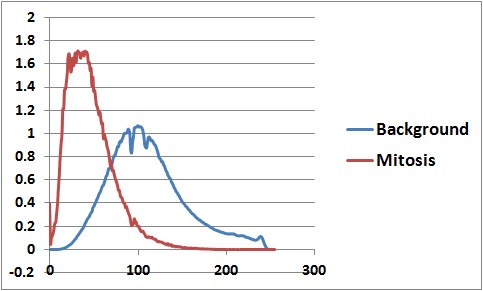}
		\caption*{g. Spectral Band 6}
	\end{subfigure}
	\begin{subfigure}[b]{0.30\textwidth}
		\includegraphics[width=\textwidth]{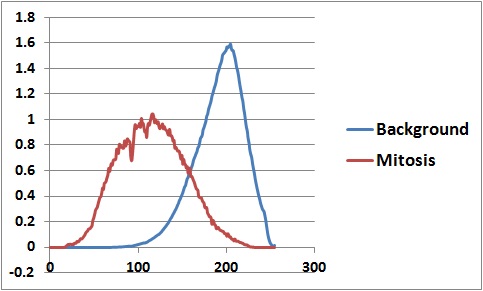}
		\caption*{h. Spectral Band 7}
	\end{subfigure}
	\begin{subfigure}[b]{0.30\textwidth}
		\includegraphics[width=\textwidth]{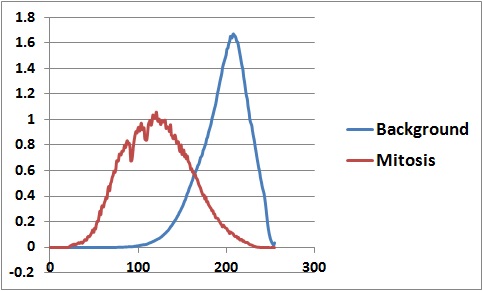}
		\caption*{i. Spectral Band 8}
	\end{subfigure}
	\begin{subfigure}[b]{0.30\textwidth}
		\includegraphics[width=\textwidth]{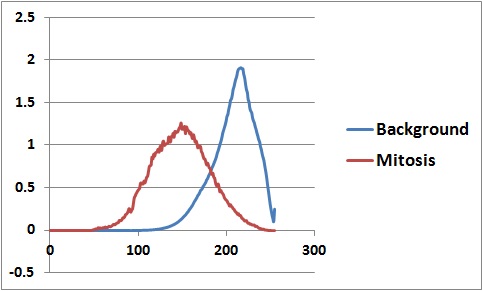}
		\caption*{j. Spectral Band 9}
	\end{subfigure}
	\caption{Histogram analysis of mitotic and background regions in 10 spectral bands.}
	\label{fig:bands_histogram}	
\end{figure}

\begin{figure}
	\centering
	\includegraphics[width=0.5\textwidth]{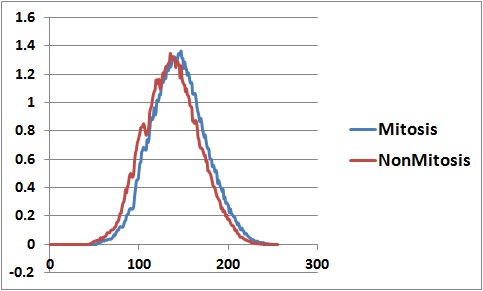}
	\caption{Histogram analysis of mitosis and non-mitosis regions in 10 spectral bands. }
	\label{fig:mitosis_nonmitosis_histogram}	
\end{figure}

Initially, we perform gradient analysis on all the focus planes. Planes with lower average gradient (i.e. out of focus) are discarded. We select focus planes 5, 6, 7, 8, 9 and 10 (total 6 focus planes). On selected focus planes, we compute the spectral responses of mitotic nuclei and background regions for all the available 10 spectral bands as represented in Fig.~\ref{fig:bands_histogram} while responses of mitotic and non mitotic nuclei are presented in Fig.~\ref{fig:mitosis_nonmitosis_histogram}. Note that spectral band 1 (white band), in nature, is different from others and might serve as reference as it contains all the information that other bands are containing, even if at a lower resolution. The peaks of the mitotic and non-mitotic nuclei are almost similar. Mitotic nuclei have different peaks than background regions. Multispectral data is thus able to differentiate between different tissue parts but spectral response of mitotic and non-mitotic nuclei is not distinguishable. The process of cell division has four different stages and each has different shape, size and textures. This motivates further spatial and morphological analysis on multispectral data to achieve reasonable classification of regions into mitotic and non-mitotic types. 

Using selected focus plane and spectral band, we eventually perform thresholding followed by morphological processing to eliminate small regions and fill holes and later select the candidates by filtering based on minimum size ($37\mu\text{m}^2$ i.e., 200 pixels) and maximum size ($1000\mu\text{m}^2$ i.e., 5405 pixels) of mitotic nuclei while keeping the number of candidates to classify lower. 

\subsubsection{Multispectral Features Computation}
Instead of single band as in \cite{boucheron2007, masood2009, wu2012, malon2013} we compute MSSF vector having intensity and textural features in all bands (10 bands). In addition, we also include morphological features (such as area, roundness, elongation, perimeter and equivalent spherical perimeter) that are computed from regions segmented during candidate detection. The morphological features reflect the phenotype information of mitotic nuclei. Using spatial information in multispectral bands, we compute five intensity-based features including mean, median, variance, kurtosis and skewness for candidate. This results in 50 multispectral intensity-based features. Haralick co-occurrence (HC) \cite{haralick1973} and run-length (RL) \cite{galloway1975} features are computed with one displacement vector in four direction $(0^o, 45^o, 90^o, 135^o)$ for all the spectral bands as in \cite{irshad2013b}. These multispectral textural features are rotationally invariant. So by making average in all four directions, eight HC and ten RL features are computed for each candidate in each single spectral band. The resulted multispectral textural features vector consists of 80 HC features and 100 RL features for each candidate. The final MSSF vector contains 235 features for each candidate. 

\subsubsection{Feature Selection and Classification}
Conceptually, a large number of descriptive features are highly desirable for classification of nuclei as mitosis or non-mitosis types. However, when using all computed features (i.e. 235 features) the classification performance is poor. Some features are irrelevant for classification and some features are redundant degrading the classification performance. The consistency subset evaluation method~\cite{liu96} is employed to select a subset of features that maximize the consistency in the class values. A projection of subset of features is performed on training dataset in order to evaluate the worth of subsets of features by the level of consistency in the class value. The consistencies of these subsets are not less than that of the full set of features. Later, we used these subsets in conjunction with a hill climbing search method, augmented with backtracking value 5, which looks for the smallest subset with consistency equal to that of the full set of features. This procedure achieved 88\% reduction in the dimensionality of features set by selecting 28 features. 

The selected features contained one morphological, 7~intensity and 20~textural features in different spectral bands except spectral band~5. The selected features set is used to train different classifiers like decision tree (Funcational Tre-FT), neural network (Multilayer Perceptron-MP), Bayesian, linear SVM (L-SVM) and non-linear SVM (NL-SVM) classifiers~\cite{weka12}. Throughout the experiments, the parameters used in FT classifier are numBoostingIterations = 15 and minNumInstance = 15, in MP are learning rate = 0.3 and momentum = 0.2, in L-SVM classifier are bias = 1, cost = 1, eps = 0.01 and kernel = L2-loss SVM and in NL-SVM classifier are kernel = rbf, degree of kernel = 3, eps = 0.001 and loss = 0.1.

%------------------------------------------------------------------------
\section{Experiment Results} \label{sec:results}
The proposed framework is evaluated on MITOS multispectral dataset \cite{mITOS2012}. The results of candidate detection and classification methods are compared with ground-truth information provided along with the dataset. The metrics used to evaluate the mitosis detection of each method include: number of true positive (TP), number of false positives (FP), number of false negative (FN), sensitivity or true positive rate (TPR), precision or positive predictive value (PPV) and F-Measure. In addition to MITOS contest metrics, the proposed framework is also evaluated with other state of the art method as 5-fold cross validation. 

\subsection{Candidate Detection in Different Spectral Bands}
\begin{figure}
	\centering
	\begin{subfigure}{0.8\textwidth}
		\includegraphics[width=\textwidth,height=50mm]{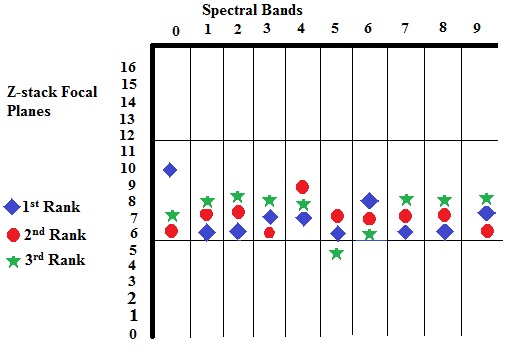}
		\caption{Top three rank focal planes.}
	\end{subfigure}
	\begin{subfigure}{0.8\textwidth}
		\includegraphics[width=\textwidth]{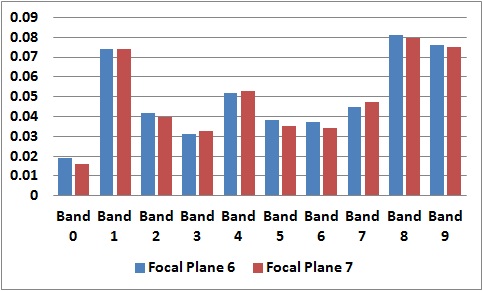}
		\caption{F-Measure in Focal planes 6 and 7.}
	\end{subfigure}
	\begin{subfigure}{0.8\textwidth}
		\includegraphics[width=\textwidth,height=50mm]{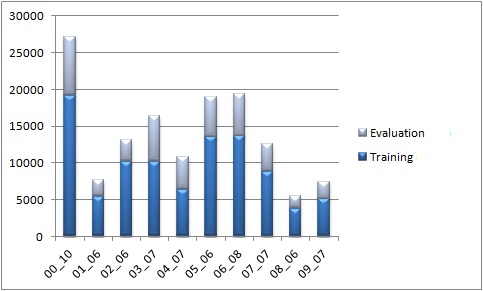}
		\caption{Candidates number in training and evaluation sets according to highest F-Measure in 10 spectral bands and selected focus planes. ($BandNumber\_FocusPlane$)}
	\end{subfigure}
	\caption{Candidate Detection Results}
	\label{fig:CandidateDetectionResults}
\end{figure}

To gain a better understanding of the relative contributions of specific spectral bands, we perform candidate detection in all available spectral bands and six selected z-stack focus planes. The results of candidate detection are ranked according to F-Measure and reported top three rank results in Fig.~\ref{fig:CandidateDetectionResults} (a). The focus planes six and seven have more information for candidate detection as compared to other focus planes and a bar graph representing candidate detection in focus planes six and seven with respect to F-Measure is shown in Fig.~\ref{fig:CandidateDetectionResults} (b). 

The important parameter that affects the classification is the unbalance training set having a large number of non-mitosis as compared to few mitosis. According to highest F-Measure in all spectral bands and focus planes, the number of detected candidates in training and evaluation sets is shown in Fig.~\ref{fig:CandidateDetectionResults} (c).  Spectral band eight in focus plane six (590-640 nm) is selected for final candidate detection as it detect less FP as compared with other spectral bands and more TP than other spectral bands as well. On training and evaluation sets, the candidate detection using spectral band 8 in focus plane 6 detects 3583 and 1655 candidates, containing 203 and 91 ground-truth (GT) mitosis from a total 224 and 98 GT mitosis, respectively. Therefore, among the entire detected candidates, there are 3380 and 1564 non-mitosis in the training and evaluation sets, respectively. The candidate detection stage generates a large number of non-mitosis and misses 21 and 7 GT mitosis, from training and evaluation sets respectively. 

\subsection{Candidate Classification}
\subsubsection{Experiment 1: Classification using separate training and evaluation sets}
In experiment 1, training set is used to train the five selected classifiers and evaluation set is used to test the classification accuracy of proposed framework as shown in Table~\ref{table:classifictionResult}. When all MSSF are used for training and evaluation, we get good TPR and F-Measure with Bayesian and FT classifiers. The MP, L-SVM and NL-SVM classifiers have few FP but also few TP and resulted low F-Measure. In case of selected MSSF, F-Measure is increase more in Bayesian classifiers with highest TPR than other classifiers. Overall, Bayesian classifier reports highest TPR (60\%) and F-Measure (57\%) while FT classifier has second highest F-Measure with few FP and high PPV.

\begin{table}
	\begin{center}
		\caption{Classification result (GT = 98) on evaluation set.}
		\label{table:classifictionResult}
	{\small
	\begin{tabular}{ p{3.5cm} c c c c c }
	\hline
	\noalign{\smallskip} 
	{\bf Experiments} & {\bf TP} & {\bf FP} & {\bf TPR} & {\bf PPV} & {\bf F-Measure}	\\
	\noalign{\smallskip}
	\hline 
	\hline
	\multicolumn{6}{c}{{\bf Using all MSSF}}\\
	L-SVM Classifier  		& 33 &  {\bf 8} & 34\% & {\bf 80\%} & 47\%\\ 
	NL-SVM Classifier 		& 40 & 28 & 41\% & 59\% & 48\%\\ 
	Bayesian Classifier     & 55 & 54 & 56\% & 50\% & 53\%\\ 
	MP Classifier     		& 36 & 15 & 37\% & 71\% & 48\%\\ 
	FT Classifier     		& 50 & 35 & 51\% & 59\% & 55\%\\ 
	\noalign{\smallskip}
	\hline 
	\noalign{\smallskip}
	\multicolumn{6}{c}{{\bf Using selected MSSF}} \\
	L-SVM Classifier  		& 35 &  9 & 36\% & {\bf 80\%} & 49\%\\ 
	NL-SVM Classifier 		& 41 & 25 & 42\% & 62\% & 50\%\\ 
	Bayesian Classifier     & {\bf 59} & 51 & {\bf 60\%} & 54\% & {\bf 57\%}\\ 
	MP Classifier     		& 33 & 13 & 34\% & 72\% & 46\%\\ 
	FT Classifier     		& 50 & 28 & 50\% & 64\% & 56\%\\ 
	\hline
	\end{tabular}
	}
	\end{center}
\end{table}

\subsubsection{Experiment 2: Classification using 5-fold cross validation}
In experiment 2, the assessment of classification performance is made using 5-fold cross validation by combining both training and evaluation set as shown in Table~\ref{table:classifictionResult2}. In case of full set of MSSF during training and evaluation, FT classifier outperforms with highest TPR and F-Measure. Overall, Bayesian classifier reported highest TPR (62\%) but more FP as well. In case of selected MSSF, as compared with L-SVM, NL-SVM and MP classifiers, FT and Bayesian classifiers report better F-Measure 54\% and 52\%, respectively.   

\begin{table}
	\begin{center}
		\caption{Classification result (GT = 322) using 5-Fold Cross Validation.}
		\label{table:classifictionResult2}
	{\small
	\begin{tabular}{ p{3.5cm} c c c c c }
	\hline
	\noalign{\smallskip} 
	{\bf Experiments} & {\bf TP} & {\bf FP} & {\bf TPR} & {\bf PPV} & {\bf F-Measure}	\\
	\noalign{\smallskip}
	\hline 
	\hline
	\multicolumn{6}{c}{{\bf Using all MSSF}}\\
	L-SVM Classifier  		& 95 & 30   & 30\% & 76\% & 43\%\\ 
	NL-SVM Classifier 		& 95 &  37  & 30\% & 72\% & 42\%\\ 
	Bayesian Classifier     & {\bf 201} & 384 & {\bf 62\%} & 34\% & 44\%\\ 
	MP Classifier     		& 132 & 90  & 41\% & 59\% & 49\%\\ 
	FT Classifier     		& 168 & 140 & 52\% & 55\% & 53\%\\ 
	\noalign{\smallskip}
	\hline 
	\noalign{\smallskip}
	\multicolumn{6}{c}{{\bf Using selected MSSF}} \\
	L-SVM Classifier  		& 96  & {\bf 24}  & 30\% & {\bf 80\%} & 43\%\\ 
	NL-SVM Classifier 		& 99  & 35  & 31\% & 74\% & 43\%\\ 
	Bayesian Classifier     & 169 & 165 & 52\% & 51\% & 52\%\\ 
	MP Classifier     		& 136 & 95  & 42\% & 59\% & 49\%\\ 
	FT Classifier     		& 141 & 62  & 44\% & 69\% & {\bf 54\%}\\ 
	\hline
	\end{tabular}
	}
	\end{center}
\end{table}

\subsubsection{Experiment 3: Classification using white spectral bands vs other multispectral bands}
To investigate the relative contribution of multispectral bands, we also perform a comparative study of mitosis classification using spatial features from white spectral band (spectral band 1) with spatial features from multispectral bands and the achieved results are shown in Table~\ref{table:classifictionResult3}. It is important to note that the multispectral bands features (MSBF) excluding white spectral band outperforms with all classifiers and reports highest F-measure 55\% with FT classifier. The classification results obtained with all five classifiers are worst using white spectral band features (WSBF). This experiment illustrates that multispectral band have much more information for mitosis classification than white bands.

\begin{table}
	\begin{center}
		\caption{Classification result (GT = 322) using 5-Fold Cross Validation.}
		\label{table:classifictionResult3}
	{\small
	\begin{tabular}{ p{3.5cm} c c c c c }
	\hline
	\noalign{\smallskip} 
	{\bf Experiments} & {\bf TP} & {\bf FP} & {\bf TPR} & {\bf PPV} & {\bf F-Measure}	\\
	\noalign{\smallskip}
	\hline 
	\hline
	\multicolumn{6}{c}{{\bf Using WSBF}} \\
	L-SVM Classifier  	& 25 		& {\bf 13}	& 8\% 			& 66\% 			& 14\%\\ 
	NL-SVM Classifier 	& 28		& 27		& 9\% 			& 51\% 			& 15\%\\ 
	Bayesian Classifier & 84 		& 236 		& 26\%			& 26\% 			& 26\%\\ 
	MP Classifier     	& 35 		& 40  		& 11\% 			& 47\% 			& 18\%\\ 
	FT Classifier     	& 35 		& 36 		& 11\% 			& 49\% 			& 18\%\\ 
	\noalign{\smallskip}
	\hline 
	\noalign{\smallskip}
	\multicolumn{6}{c}{{\bf Using MSBF}} \\
	L-SVM Classifier  	& 82  		& 57    	& 25\% 			& 59\% 			& 36\%\\ 
	NL-SVM Classifier 	& 83  		& 78  		& 26\% 			& 52\% 			& 34\%\\ 
	Bayesian Classifier & {\bf 197}	& 615 		& {\bf 61\%}	& 24\% 			& 35\%\\ 
	MP Classifier     	& 109		& 43  		& 34\% 			& {\bf 72\%}	& 46\%\\ 
	FT Classifier     	& 140  		& 84  		& 43\% 			& 63\% 			& {\bf 51\%}\\ 
	\hline
	\end{tabular}
	}
	\end{center}
\end{table}
 
\subsubsection{Experiment 4: Multispectral texture features vs multispectral intensity features}
In this experiment, we explore the impact of multispectral texture features (MSTF) and multispectral intensity features (MSIF) on the mitosis classification. The results are shown in Table~\ref{table:classifictionResult4}. In case of MSIF, we get high F-Measure 55\% using MP classifier. The highest F-Measure (56\%) is achieved with MSTF with FT classifier. These results illustrate that MSTF have more information for mitosis classification as compared MSIT.
  
\begin{table}
	\begin{center}
		\caption{Classification result (GT = 322) using 5-Fold Cross Validation.}
		\label{table:classifictionResult4}
	{\small
	\begin{tabular}{ p{3.5cm} c c c c c }
	\hline
	\noalign{\smallskip} 
	{\bf Experiments} & {\bf TP} & {\bf FP} & {\bf TPR} & {\bf PPV} & {\bf F-Measure}	\\
	\noalign{\smallskip}
	\hline 
	\hline
	\multicolumn{6}{c}{{\bf Using MSIF}} \\
	L-SVM Classifier  	& 81  		& {\bf 38}	& 25\% 			& {\bf 68\%}	& 37\%\\ 
	NL-SVM Classifier 	& 86 		& 63   		& 27\% 			& 58\% 			& 37\%\\ 
	Bayesian Classifier & 154 		& 278		& 48\%			& 36\% 			& 41\%\\ 
	MP Classifier     	& 156 		& 91  		& 48\% 			& 63\% 			& 55\%\\ 
	FT Classifier     	& 140 		& 83  		& 43\% 			& 63\% 			& 51\%\\ 
	\noalign{\smallskip}
	\hline 
	\noalign{\smallskip}
	\multicolumn{6}{c}{{\bf Using MSTF}} \\
	L-SVM Classifier  	& 84 		& 41   		& 26\% 			& 67\% 			& 38\%\\ 
	NL-SVM Classifier 	& 88 		& 69   		& 27\% 			& 56\% 			& 37\%\\ 
	Bayesian Classifier & {\bf 200}	& 553  		& {\bf 62\%}	& 27\% 			& 37\%\\ 
	MP Classifier     	& 137		& 65  		& 43\% 			& {\bf 68\%}	& 52\%\\ 
	FT Classifier     	& 167 		& 105 		& 52\% 			& 61\% 			& {\bf 56\%}\\ 
	\hline
	\end{tabular}
	}
	\end{center}
\end{table}
%------------------------------------------------------------------------

\section{Discussion} \label{sec:Discussion}
The results seem to indicate the best scores are achieved using the same focus plane, or couple of focus planes, across bands. Each of those planes is exhibiting maximum gradient intensity above the rest of the focus planes. It means that in further experiments we could locate the best focus planes first, and then apply our framework on that limited dataset. In other words, finding the best focus plane and finding the best bands are separable problems.

The best F-Measure for candidate detection is achieved on spectral bands 8, 9 and 1 respectively. The fact that candidate detection using bands 8 and 9, the proposed framework achieves better results than when using the full spectrum (band 1) supports the claim that MSI improves the accuracy of the framework. As bands 8 and 9 actually overlap in terms of spectrum, it would be interesting to try to apply spectral unmixing between bands 8 and 9 to see if it can further improve the results. The results illustrate clearly the improved accuracy resulting of the selection process.

We use the analysis of different subset of multispectral features as complement to the analysis of performance on MSSF. Specifically, 5-fold cross validation is employed to perform comprehensive analysis of proposed framework. In order to study the MSSF for classification of mitosis figures, we perform features selection in different spectral bands by studying which spectral bands have redundant and irrelevant information for mitosis classification. Spectral band 5 is irrelevant for mitosis discrimination. Most of selected MSSF belong to spectral band 7 and 8. Most of the selected features are texture in different spectral bands, especially spectral bands 7 and 8, which strengthening the importance of multispectral spatial analysis. 

Specifically we have shown in experiment 3 that most of the information is present in multispectral band excluding white spectral band that is not suitable for mitosis classification. According to experiments 4, multispectral texture features are more helpful for mitosis discrimination as compared to multispectral intensity features. 

\begin{figure}[t]
	\centering
	\includegraphics[width=0.9\textwidth]{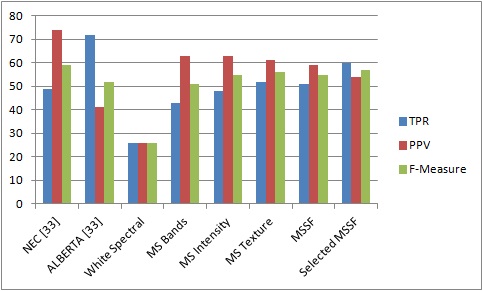}
	\caption{Comparison of proposed framework results with MITOS contest result.}
	\label{fig:mitosComparison}	
\end{figure}

\begin{figure}[t]
	\centering
	\includegraphics[width=\textwidth]{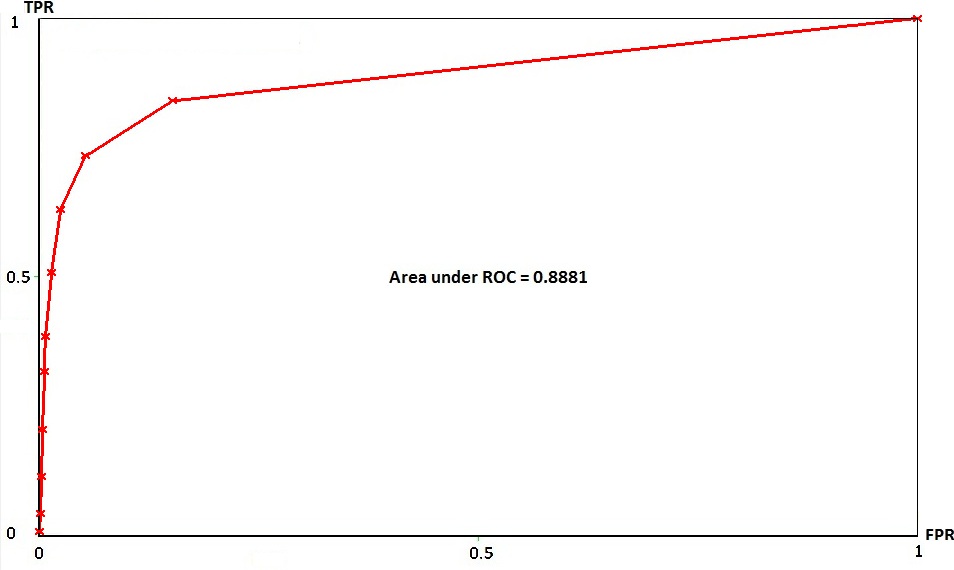}
	\caption{The ROC curve using selected MSSF with FT classifier.}
	\label{fig:roc}	
\end{figure}

Separate training and evaluation sets as provided with MITOS dataset \cite{mITOS2012} have been used for training and evaluation of proposed framework. To put the quality of this result in perspective, in comparison with MITOS contest result \cite{mITOS2012} proposed method with selected MSSF and Bayesian classifier managed to achieve second highest TPR and F-Measure. The comparison of proposed framework results with MITOS contest results are shown in Fig.~\ref{fig:mitosComparison}. Fig.~\ref{fig:roc} illustrates the ROC curve obtained with selected MSSF with FT classifier. This clearly demonstrates that our new proposed framework results in an improved ability to distinguish mitosis from other objects. The proposed framework should be considered as the state of the art.

\section{Conclusion} \label{sec:conclusion}
An automated mitosis detection framework for breast cancer MSI based on multispectral spatial features has been proposed. Initially, candidate detection is performed in selected spectral band and z-stack focus plane. Then, we compute multispectral spatial features for each candidate, a highly efficient model for capturing texture features for region (nuclei) discrimination. The proposed framework reaches the best levels of the MITOS contest results. Both features set (MSIF \& MSTF) shown similar classification performance for mitosis. In future work, we plan to investigate unmixing of bands and/or better selection of (sets of) bands of interest. The pre-selection of the focus plane (or volumes) is also of great importance to reduce the complexity of the dataset and improve the actual performances to reach clinical operational acceptance expected by our professional consortia.

%------------------------------------------------------------------------
\section*{Acknowledgement}
This work was partially supported by the French National Research Agency ANR, project MICO under reference ANR-10-TECS-015.

\bibliographystyle{elsarticle-num}

\end{document}